# Spiralet Sparse Representation
# Working Paper WP-RFM-14-01, (version: 140409)

Reza Farrahi Moghaddam[a,*], Mohamed Cheriet[a]

[a]*Synchromedia Lab, École de technologie supérieure (ETS), University of Quebec (UduQ), Montreal, QC, Canada*

**Abstract**

This is the first report on Working Paper WP-RFM-14-01. The potential and capability of sparse representations is well-known. However, their (multivariate variable) vectorial form, which is completely fine in many fields and disciplines, results in removal and filtering of important "spatial" relations that are implicitly carried by two-dimensional [or multi-dimensional] objects, such as images. In this paper, a new approach, called *spiralet sparse representation*, is proposed in order to develop an augmented representation and therefore a modified sparse representation and theory, which is capable to preserve the data associated to the spatial relations.

## 1. Introduction

Sparse representations relax the orthogonality constraint in terms of the representation's bases. This is of great advantage with respect to increasing the redundancy and also make the representation less sensitive to its bases. However, it comes with the cost of losing the rich spatial, [spectral, among others] relations and interrelations of the input data, especially in the case of multi-dimensional objects such as images. A possible solution, which is proposed here, is to remove the vector-form (multivariate variable) condition of traditional sparse and other similar representations.

*Corresponding author: Reza Farrahi Moghaddam
  *Email addresses:* imriss@ieee.org (Reza Farrahi Moghaddam),
mohamed.cheriet@etsmtl.ca (Mohamed Cheriet)
  *URL:* http://ca.linkedin.com/in/rezafm (Reza Farrahi Moghaddam)

## 2. General Notation

In this section, the notation used to express the traditional sparse representation in the next section is presented.

- $X$: The observed image (or patch):

$$X = (x_{i,j})_{i=1,j=1}^{n,m}.$$

- $Y$: The *equivalent* vectorial signal associated to $X$ (in the observation space $\mathcal{O}$):

$$Y = (y_k)_{k=1}^{nm},$$

where $y_k = x_{i_k, j_k}$ using a column (or row)-based order: $j_k = [k/n] + 1$, $i_k = k - m * j_k$.

- $D$: The dictionary of the model:

$$D = (Y_\eta)_{\eta=1}^{d},$$

where $Y_\eta$ is one of $\eta$ selected signal that represent the dictionary's atoms.

## 3. The Traditional Sparse Representation

Starting from an observed image $X$, the observed signal $Y$ can be seen as an equivalent multivariate variable (vector). This implicitly removes the spatial relations carried by $X$, and can be seen as the first *flattening* change.

The traditional model is described as:

$$Y = D\Theta := \sum_{\eta} Y_\eta \Theta_\eta, \tag{1}$$

where $\Theta$ is the [sparse] representation of $Y$ in terms of the dictionary $D$. This linear relation is the second *flattening* change.

The sparseness is enforced by controlling non-zero elements of $\Theta$. For example, [2, 6, 7],

$$|\Theta|_0 \leq k$$

or

$$\Theta = \operatorname*{argmin}_{\Theta', \forall Y \in \mathcal{O}} \left\lVert \Theta' \right\rVert_0.$$

This is usually expressed in the form of an optimization problem. However, because of computational concerns, other norms such as $\lVert \cdot \rVert_1$ are considered.

The main challenge is on how to avoid the loss of the *spatial relations*. It can be argued that the partial-order representations, such as the Spiral Representation, preserves the spatial connectivity at least at an one-pixel scale [5]. In this work, we generalize this concept into the *Spiralet Transform* that represent an image or a patch as a whole while preserving its spatial relations.

## 4. The Notation of the Proposed *Spiralet Sparse* Representation

In this section, additional notations used to describe the proposed spiralet sparse representation are provided.

- $\mathcal{X}^{[l,s]}$: The *Spiralet Transform* of order $l$ and radius $s$ of an image $X$:

$$\mathcal{X}^{[l,s]} = \left(\S_{i,j;\omega_1,\cdots,\omega_l}^{[l,s]}\right)_{i,j;\omega_1,\cdots,\omega_l},$$

where $\S_{i,j;\omega_1,\cdots,\omega_l}^{[l,s]} = S\left(i, j, \mathcal{X}^{[l-1,s]}\right)$, and $S\left(\cdot\right)$ is the *spiralet transform function*. For $l = 0$, $\S_{i,j}^{[0,\cdot]} = x_{i,j}$, i.e., $\mathcal{X}^{[0,\cdot]} = X$. The spiralet has $l$ number of extra dimensions in its representation. In this work, we assume $l = 1$ and $s = 1$, i.e., the first-order spiralets of radius one. This allows us to represent the spiralet $\mathcal{X}$ as a rectangular cube with a depth dimensions along the index of $S(x_{i,j})$, we call $\omega$, $\omega = 1, \cdots, 9$. It can be easily observed that:

$$\mathcal{X}^{[1,1]} = \left(\S_{i,j;\omega}^{[1,1]}\right)_{i,j;\omega}.$$

- The length of $S$'s elements in one level expansion is denoted $\bar{s}$: For $s = 1$, $\bar{s} = 9$. For $s = 2$, $\bar{s} = 25$.

- For $l = 1$ and $s = 1$, $S$ is defined:

$$\begin{aligned}\S_{i,j;\cdot}^{[1,1]} = S\left(i, j, \mathcal{X}^{[0,1]}\right) = &\left(S^{[1]}(x_{i,j}, x_{i,j}), S^{[1]}(x_{i,j}, x_{i-1,j}), S^{[1]}(x_{i,j}, x_{i-1,j+1}),\right.\\ &S^{[1]}(x_{i,j}, x_{i,j+1}), S^{[1]}(x_{i,j}, x_{i+1,j+1}), S^{[1]}(x_{i,j}, x_{i+1,j}),\\ &\left.S^{[1]}(x_{i,j}, x_{i+1,j-1}), S^{[1]}(x_{i,j}, x_{i,j-1}), S^{[1]}(x_{i,j}, x_{i-1,j-1})\right),\end{aligned}$$

Figure 1: A typical expansion of the spiral representation [5].

where $S^{[1]}(\alpha, \beta)$ is a nonlinear function defined:

$$S^{[1]}(\alpha, \beta) = \Big((1 - \text{SIM}(\alpha, \beta))\alpha + (\text{SIM}(\alpha, \beta))\beta\Big),$$

and $\text{SIM}(\alpha, \beta)$ is a *similarity function*. In examples here, we use:

$$\text{SIM}(\alpha, \beta) = \exp\left(-\left\|\alpha - \beta\right\|/h\right), \qquad (2)$$

where $h$ is a bandwidth parameter.

- For $s > 1$, the definition can be extended as in [5] to include other neighboring pixels in a spiral way (See Figure 1). The pixels at radii $s = 1$ and $s = 2$ are shown in blue and green.

- $\mathcal{Y}$: The *Spiralet Signal* associated to image $X$ (the $l = 1$ and $s = 1$ indices are removed for simplicity):

$$\mathcal{Y} = (\dagger_{k,\omega})_{k,\omega} = (\S_{i_k, j_k; \omega})_{k=1, \omega=1}^{nm, \bar{s}},$$

where $\S_{i_k, j_k; \omega}$ is an element of the first-level spiralet $\mathcal{X}$ at position $k$

(or, pixel $(i_k, j_k)$), and $\bar{s}$ is the length of each spiralet. For first-order of radius one, we have: $\bar{s} = 9$. $\mathcal{Y}$'s spiralet nature partially preserves the spatial relations, and therefore is capable to mitigate and dampen the first flattening effect.

- $\mathcal{Y}$ can be imagined as a matrix with vertical dimension along $k$ and *in-depth* dimension along $\omega$.

- $\mathcal{D}$: The generalized dictionary is defined similar to $\mathcal{Y}$:

$$\mathcal{D} = \mathcal{D}_{\eta,k,l} = (\mathcal{Y}_\eta)_{\eta=1}^d.$$

$\mathcal{D}$ can be imagined a rectangular cube where the depth represents the spiralet generalization, i.e., the $\omega$ index. It resembles putting spiralet signals $\mathcal{Y}_\eta$s as slices side by side along the $\eta$ index.

## 5. The Proposed Model of the *Spiralet Sparse* Representation

The new model is consequentially defined:

$$\mathcal{Y} = \mathcal{D} \otimes_\eta \Theta := \sum_\eta \mathcal{D}_{\eta,k,l}\Theta_\eta, \tag{3}$$

which is a linear model in the spiralets space, and $\otimes_\eta$ denotes a summation on $\eta$ index. The spiralet model has $s$ times more equations. Therefore, the dictionary should also expanded accordingly in order to preserve an acceptable level of freedom and variability.

In addition, equivalently in a weak form, the model could be expressed as follows in terms of the original signal:

$$Y = S^{[-1]} \left( \mathcal{D} \otimes_\eta \Theta \right) := S^{[-1]} \left( \sum_\eta \mathcal{D}_{\eta,k,l} \Theta_\eta \right), \tag{4}$$

where $S^{[-1]}$ is an *activation* function. For the first-level spiralets of radius one, an example of an *isotropic* $S^{[-1]}$ could be:

$$S^{[-1]}\left(\mathcal{Y}'\right) = \left\{ \left(1, \exp\left(\frac{-1}{h}\right), \cdots, \exp\left(\frac{-1}{h}\right)\right)^{\mathrm{T}} \times \left(\dagger'_{k,\cdot}\right) \right\}, \tag{5}$$

where $\mathcal{Y}' = \left(\dagger'_{k,\omega}\right)_{k,l}$ is a spiralet signal that equates the $\mathcal{D} \otimes_\eta \Theta$ in the model in the spiralet space.

## 6. An Illustrative Example

We consider the simplest image in this example: $X = (x_{1,1}, x_{2,1})$, i.e., $n = 2$, $m = 1$. Also, Let's assume $x_{1,1} = 0.3$ and $x_{2,1} = 0.6$. We assume a "frozen" dictionary of three $(\eta = 1, \cdots, 3)$ dictionary atoms: $Y_1 = (1, 0)$, $Y_2 = (0, 1)$, and $Y_3 = (1, 1)$.

A *traditional* sparse solution $\Theta_t$ according to traditional sparse problem,

$$\Theta_t = \operatorname*{argmin}_{\Theta'_s} ||\Theta'_t||_0,$$

would be:

$$\Theta_t = (0, 0.3, 0.3)$$

with $||\Theta_t||_0 = 2$.

In terms of Spiralet transform,

$$\mathcal{X} = \begin{pmatrix} 0.3, & 0.41 \\ 0.6, & 0.49 \end{pmatrix},$$

when $h = 0.3$ is considered. Also,

$$\mathcal{Y}_1 = \begin{pmatrix} 1, & 0.96 \\ 0, & 0.04 \end{pmatrix},$$

$$\mathcal{Y}_2 = \begin{pmatrix} 0, & 0.04 \\ 1, & 0.96 \end{pmatrix},$$

and

$$\mathcal{Y}_3 = \begin{pmatrix} 1, & 1 \\ 1, & 1 \end{pmatrix}.$$

The spiralet dictionary, $\mathcal{D}$, is simply built by stacking the $\mathcal{Y}_\eta$s side by side.

A *spiralet* sparse solution $\Theta_s$ according to spiralet sparse problem,

$$\Theta_s = \underset{\Theta'_s}{\operatorname{argmin}} \left\lVert \Theta'_s \right\rVert_0,$$

would be:

$$\Theta_s = (0, 0.2, 0.35)$$

with $\lVert \Theta_s \rVert_0 = 2$. The spiralet sparse has a small intrinsic error even without any filter: $\lVert e_s \rVert_2 = 0.071$.

Although the dictionary is set fixed in this example, in order to show the power and potential of the spiralet transforms and its sparse representations, we evaluate the resiliency of the two representations with respect to some filters:

### 6.1. Over-sparseness filter

Let's impose a filter that enforces $||\theta|| = 1$.

For traditional sparse case, we have: $\theta''_t = c(0, 0, 0.3)$ after applying this filter, and corresponding error would be $||e''_t||_2 = 0.3$. In contrast, for Spiralet sparse case, we have: $\theta''_s = c(0, 0, 0.35)$, and corresponding error would be $||e''_s||_2 = 0.255$, which shows **less** error and sensitivity.

## 7. The Discussions

It is worth noting that in Equations (2) and (5), the parameter $h$ behaves similar to the bandwidth of NLM and NLPM [1, 3, 4]. Also, the activation function $S^{[-1]}$ can be seen very close to that used in Neural Networks (NNs). Therefore, the spiralet representation and modeling can be imagined as a generalization to traditional NNs that preserves the spatial relations by adding another dimension [along the spiralet index]. This brings up the challenge of how to add abstraction and deep learning to this new representation. Finally, it is worth mentioning that the spiralet transformation is not limited to only sparse representation, and it can be also used to make other representations spatial-ware.

## 8. The Conclusions

A new sparse representation for images and other multi-dimensional objects is proposed based on the spiral representation. The new representation, called spiralet sparse representation, has the capability to preserve the spatial relations in the form of extra dimensions of the representation. A simple illustrative example has been provided. The formulation and use cases will be expanded.

# References

[1] A. Buades, B. Coll, and J.-M. Morel. A non-local algorithm for image denoising. In B. Coll, editor, *CVPR'05*, volume 2, pages 60–65 vol. 2, 2005.

[2] M. Elad. Sparse and redundant representation modeling–what next? *IEEE Signal Processing Letters*, 19(12):922–928, 2012.

[3] Reza Farrahi Moghaddam and Mohamed Cheriet. Beyond pixels and regions: A non-local patch means (NLPM) method for content-level restoration, enhancement, and reconstruction of degraded document images. *Pattern Recognition*, 44(2):363–374, 2011.

[4] Reza Farrahi Moghaddam and Mohamed Cheriet. Real-time knowledge-based processing of images: Application of the online NLPM method to perceptual visual analysis. *IEEE Transactions on Image Processing*, 21(8):3390–3404, 2012.

[5] Reza Farrahi Moghaddam, Fereydoun Farrahi Moghaddam, and Mohamed Cheriet. A new framework based on signature patches, micro registration, and sparse representation for optical text recognition. In *ISSPA'12*, pages 1259–1265, 2012.

[6] P. Sprechmann and G. Sapiro. Dictionary learning and sparse coding for unsupervised clustering. In *ICASSP'10*, pages 2042–2045, 2010.

[7] J. Wright, Yi Ma, J. Mairal, G. Sapiro, T.S. Huang, and Shuicheng Yan. Sparse representation for computer vision and pattern recognition. *Proceedings of the IEEE*, 98(6):1031–1044, 2010.